# Generative Multiple-Instance Learning Models
# For Quantitative Electromyography

**Tameem Adel**†   **Ruth Urner**‡   **Benn Smith**\*   **Daniel Stashuk**†   **Daniel J. Lizotte**‡

†Systems Design Engineering
‡David R. Cheriton School of Computer Science
University of Waterloo, Waterloo, ON, Canada

\*Department of Neurology
Mayo Clinic, Scottsdale, AZ, United States

## Abstract

We present a comprehensive study of the use of generative modeling approaches for Multiple-Instance Learning (MIL) problems. In MIL a learner receives training instances grouped together into bags with labels for the bags only (which might not be correct for the comprised instances). Our work was motivated by the task of facilitating the diagnosis of neuromuscular disorders using sets of motor unit potential trains (MUPTs) detected within a muscle which can be cast as a MIL problem. Our approach leads to a state-of-the-art solution to the problem of muscle classification. By introducing and analyzing generative models for MIL in a general framework and examining a variety of model structures and components, our work also serves as a methodological guide to modelling MIL tasks. We evaluate our proposed methods both on MUPT datasets and on the MUSK1 dataset, one of the most widely used benchmarks for MIL.

## 1   Introduction

In Multiple-Instance Learning (MIL), training instances are grouped together in *bags* which have labels. Each *instance* in a bag has a label that may be different from that of the bag, but instance labels are not observed; only the label of the bag is available for learning. The MIL framework was first introduced by Dietterich et al. [1997] for a problem in a medical (pharmaceutical) domain. Their task was to predict the binding properties of molecules, which depend on the shape of the molecule. However, a molecule can take on several shapes. Thus, each molecule is represented as a bag of instances, where each instance represents a shape the molecule can take on. If none of the possible shapes enable binding, the bag (molecule) gets a negative label. But as soon as one shape allows for binding, the bag is labeled positive. The MUSK dataset from this problem has remained one of the most widely used benchmark datasets for MIL tasks.

Following the introduction of the framework, various problems have been expressed as MIL. MIL approaches have, for example, been employed for content-based image retrieval [Maron and Ratan, 1998, Zhang et al., 2002], text classification [Settles et al., 2007, Andrews et al., 2002b], protein identification [Tao et al., 2004], music information retrieval [Mandel and Ellis, 2008] and activity recognition [Stikic and Schiele, 2009]. In medical domains, prediction problems often naturally occur as MIL tasks. One example is the original MUSK prediction task; Dundar et al. [2008] also show that learning problems for computer-aided detection applications can often be considered as MIL problems.

Our work was motivated by an application which uses quantitative analysis of clinically detected electromyographic (EMG) signals to assist with the diagnosis of certain neuromuscular disorders. The diagnosis of a neuromuscular disorder often requires the characterization of several individual muscles. A muscle characterization, in turn, is based on characterizing a sampling of its motor units (MUs). A motor unit potential train (MUPT) created by a MU and extracted from a needle-detected EMG signal can be used to obtain a characterization of the MU (see Section 2 for details). The classification of a muscle based on the set of MUPTs representing a sampling of its MUs can therefore be formulated as a MIL problem wherein each muscle is a bag and each MU of a muscle is an instance of that bag. We propose that *generative* modeling approaches—models that describe the full joint distribution of the data—are useful and effective for data that naturally occurs in MIL form, such as muscle classification based on a set of MUPTs. Predicting with a generative model is particularly suitable for medical domains for several reasons: Generative models allow for expert domain knowledge to be incorporated in an intuitive way, which leads to good inductive bias in the modeling assumptions. As we will demonstrate, a model with good inductive bias (elicited from experts in biomedicine) can result in highly accurate predictions even on the basis of a relatively small training set. Most importantly, they yield not only a classification tool, but a simulation tool for the problem domain. In our setting, such a simulator provides a stepping stone toward a more sophisticated system that not only helps with the diagnosis of neuromuscular disorders but also provides a measure of their severity.

Our contributions are two-fold: First, we provide a state-of-the-art solution to the problem of muscle classification. We show that modeling the muscle as a two-stage generative model (according to the way its MUPTs are actually generated) significantly improves classification accuracy over previous strategies for this task both at the instance and bag level [Adel et al., 2012]. Second, we introduce a general framework and provide intuition and guidelines for applying generative models to MIL problems. Generative models have only recently been successfully applied to MIL tasks [Yang et al., 2009, Foulds and Smyth, 2011]; these can be viewed as special cases of our framework. We compare different possible model structures for MIL generative models both conceptually and experimentally, and we discuss the impact of their differing conditional independence structures and parametric modeling assumptions. We suggest several possible implementations for these structures and validate the proposed methods both on the MUPT and MUSK data. Because we examine a variety of model structures and components (not just those appropriate for muscle classification) our work also serves as a methodological guide to modelling MIL tasks.

## 2 Muscle Classification using QEMG

Quantitative electromyography (QEMG) is a method used to help diagnose neuromuscular disorders by quantitatively analyzing EMG signals detected during a slight to moderate level, voluntary muscle contraction using an inserted needle electrode. A muscle is comprised of several MUs which are repeatedly activated during muscle contraction. A motor unit consists of a group of muscle fibers and the $\alpha$ motor neuron that activates those fibers. The voltage signal detected by an electrode created by the activation of the fibers of a motor unit is called a motor unit potential (MUP). The train of MUPs created by the repeated activity of a MU is called a MUPT. Each EMG signal is thus a compound signal that represents the sum of the MUPTs of all active motor units. For analysis purposes, an EMG signal is decomposed into its constituent MUPTs using a state-of-the-art pattern recognition based decomposition system (see also the work of Adel et al. [2012] and Farkas et al. [2010] for more details). Figure 1 illustrates this MUPT extraction process. Usually, 4 to 6 EMG signals, each detected with the needle at a distinct location in the muscle, are recorded and decomposed to obtain a representative set of MUPTs of sufficient size (15-25).

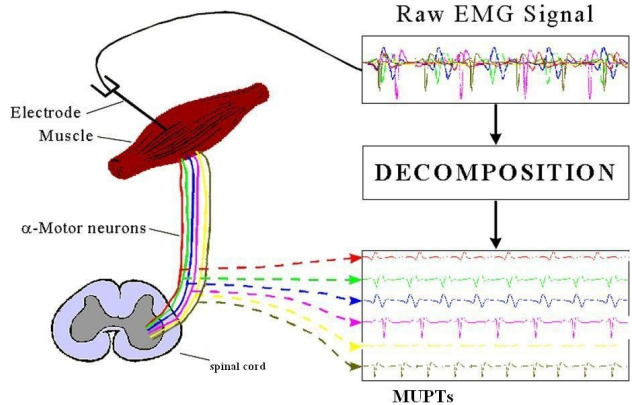

Figure 1: MUPT extraction via EMG Signal Decomposition. Derived from Basmajian and Luca [1985].

Muscles are classified as either *normal*, *myopathic* or *neurogenic* based on clinical expertise. Initially, MUPTs are labeled normal, myopathic or neurogenic on the basis of being detected in a normal, myopathic or neurogenic muscle. For normal muscles, this is largely correct as it is unlikely that a motor unit of a normal muscle would generate a disordered (myopathic or neurogenic) MUPT. However, passing on the muscle label to each MUPT is not accurate for disordered muscles: myopathic and neurogenic muscles commonly have some normal MUs, and thus produce some normal MUPTs. Thus, labelling all MUPTs in a disordered muscle as disordered is incorrect.

Classifying a muscle as normal, myopathic or neurogenic can be posed as a MIL problem in a straightforward manner: In this task, a bag corresponds to the muscle that produces the MUPT instances, with each instance representing a sampled MU within that muscle. The features of an instance correspond to the features used to represent the MUPT of the MU. The instances of a normal bag are all normal, while neurogenic and myopathic bags might contain both normal and neurogenic or myopathic instances, respectively. It is exceedingly unlikely that a neurogenic (resp. myopathic) disordered muscle contains/generates a myopathic (resp. neurogenic) MU/MUPT. While it is possible for a domain expert to manually classify individual MUPTs, this task is time-consuming. Clinical experts therefore typically provide only the diagnosis for the whole muscle, which gives the bag label. Thus a learner must learn how to classify new MUPTs and muscles from a training set providing only muscle labels.

The machine learning techniques implemented in this work are known as *quantitative EMG techniques*. The main goal of quantitative EMG techniques is to extract suitable information from detected EMG signals and then interpret this information to assist with the diagnosis of their respective muscles. It is also desirable that quantitative EMG analysis provides a measure related to the severity of the predicted disorder [Pino, 2008]. Two muscles can both be myopathic, but one may be mildly myopathic while the other is severely myopathic; it is hypothesized that severe cases are indicated by increased numbers of MUPTs from within their class as well as that some of the MUPTs may be outliers within their class. By identifying when an instance is atypical within its class label, in future our generative model may be used for estimating the severity of muscular disorders. Our ultimate goal is to build a clinical decision support system that assists with the diagnosis of muscles (both in terms of classification and assessment of severity) by inspecting sets of MUPTs extracted from EMG signals.

## 3 Related work

The last decade and a half has seen the development of a large body of work on MIL, both in terms of theoretical analysis and the development of practical algorithms for various application areas as we mentioned in the introduction.

Dietterich et al. [1997] suggest several algorithms for learning axis-aligned rectangles for the original MIL problem on the MUSK data. Maron and Lozano-Pérez [1998] introduced the diverse density (DD) algorithm, a paradigm for MIL that, similar to the axis-aligned rectangle learning approach, assumes that there is a specific region of positive instances to be identified in the feature space. The algorithm has been further developed to EM-DD by Zhang and Goldman [2001]. This is one of the most successful approaches for MIL and we discuss how it relates to our framework in Section 4.2. Wang and Zucker [2000] adapted Nearest Neighbor learning to MIL. Several studies present kernels to use Support Vector Machines on MIL problems [Gärtner et al., 2002, Andrews et al., 2002a, Tao et al., 2004], or adaptations of boosting [Andrews and Hofmann, 2003, Xu and Frank, 2004, Viola et al., 2005] and, more recently, incorporate methods from semi-supervised or active learning into the MIL setting [Rahmani and Goldman, 2006, Zhou and Xu, 2007, Settles et al., 2007]. The original bag labeling rule (where the label of the bag is the logical OR of its instances as in the MUSK data) has been modified and generalized to apply to other areas (Foulds and Frank [2010] provide an overview).

Long and Tan [1998] analyze the original problem of learning axis-aligned rectangles from MIL data in the Probably Approximately Correct (PAC) learning framework. This set-up assumes independence of the instances that occur together in a bag and the goal is to learn a low-error instance level classifier. Blum and Kalai [1998] show that this framework is equivalent to PAC-learning with one-sided noise, a problem that has recently been analyzed in Simon [2012]. However, in most MIL problems, it is not appropriate to assume that instances occurring together in a bag are conditionally independent and the goal is to learn a bag-level classifier rather than an instance-level classifier. Sabato et al. [2010] provides a comprehensive study with upper and lower bounds on the sample complexity of the bag-level learning problem without the independence assumption. Diochnos et al. [2012] tighten some of those lower bounds.

Generative model approaches have only rather recently been introduced to the MIL setting [de Freitas and Kück, 2005, Yang et al., 2009, Foulds and Smyth, 2011]. The former two studies suggest more complex model structures for modifications of the MIL problem. The work of Foulds and Smyth [2011] fits into our framework with specific choices for the model components. We discuss the modeling choices that were made in Foulds and Smyth [2011] and Yang et al. [2009] in the context of our framework below.

## 4 Generative Models for MIL

We denote random variables in upper case, and their realizations in lower case. Let $t$ be the number of possible bag/instance labels. Let $B \in \{1, 2, ..., t\}$ represent a bag's label, and let $I_j \in \{1, 2, ..., t\}$ represent the label of the $j$th instance belonging to the bag. The number of instances in the bag is denoted by $m$; we refer to the $m$ instance labels together as the vector $\vec{I}$. Let $\vec{F}_j \in \mathbb{R}^p$ be the $p$-dimensional feature vector belonging to the $j$th instance. We index elements of a vector with a square-bracketed subscript, so $\vec{F}_{j[k]}$ is the $k$th element of the observed feature vector of the $j$th instance in a bag. In our models, marginal and conditional distributions involving only $I_j$ and $\vec{F}_j$ are the same for all $j$, so we refer to a "generic" instance label as $I$ and a "generic" feature vector as $\vec{F}$.

Most MIL work to date considers binary labels, i.e. $t = 2$. In our muscle classification problem $t = 3$, since a bag or instance can be either *normal* ($B = 1$), *myopathic* ($B = 2$), or *neurogenic* ($B = 3$). Fur-thermore, in our problem, a bag may only generate a *compatible* instance label: We call the value of a single instance label $i_j$ *compatible* with a bag label $b$ iff $i_j \in \{1\} \cup \{b\}$. We call a joint labelling $\vec{i} = i_1, ..., i_m$ *feasible* iff $\exists b(\forall j, i_j \in \{1\} \cup \{b\})$.

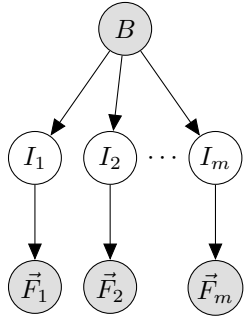

Figure 2: The BIF Model Structure. Parameters for $P(I_j|B)$ and for $P(\vec{F}_j|I_j)$ are tied across $j$.

We begin by discussing possible Bayes net structures for our MIL model in Section 4.1. We then discuss possible modelling choices for marginal and conditional distributions in Section 4.2.

## 4.1 Model Structures

The main inductive bias that we retain from the original MIL formulation is the assumption that the bag label is conditionally independent of the feature vectors given their corresponding instance labels. This is implied by the assumption that the feature distribution of normal instances in a normal bag should be the same as the feature distribution of normal instances in an abnormal bag. This restricts us to three possible Bayes net structures presented below, two of which we will use as candidate structures for our generative model. For completeness, we also discuss a fourth structure that does not satisfy the conditional independence assumption. We name the structures based on the partial order in which the variables appear in the graph. Alpaydin [2010] gives a concise overview of directed graphical models; Koller and Friedman [2009] give a comprehensive treatment.

### 4.1.1 BIF: $B \rightarrow I \rightarrow \vec{F}\ _m$

This structure best represents the generative process underlying our MUPT data. Under this structure, the bag (muscle) generates its $m$ instances (MUPTs) independently given its label. Each instance in turn generates its own feature vector given *its* label, but independent of the bag label and independent of the other instances and features in the bag. The structure of this model is given concisely by the plate diagram $B \rightarrow I \rightarrow \vec{F}\ _m$ for which we give the expanded version in Figure 2. The other model structures we consider are the same except for the directions of the edges. We will see that the choice of the edge directions has important consequences.

For this structure, we must learn $P(B)$ from observed data, and learn $P(I|B)$ and $P(\vec{F}|I)$ using a hidden variable method like EM. Constraints on $P(I|B)$ are simple to encode; to ensure that a bag can *never* generate an incompatible instance, we can restrict the values in the conditional probability table of $I|B$ by requiring $P(I = i|B = b) = 0$ for all $i \notin \{1\} \cup \{b\}$. It follows that sets of instance labels drawn from this model are always compatible with the bag label. Furthermore, we can easily impose a Dirichlet prior on the proportion of instance labels that match the bag label while obeying these constraints. Since we have continuous features, $P(\vec{F}|I)$ is modeled using density estimation.

The main departure this model makes from other probabilistic models for MIL is that it assumes that the bag label is the cause of the instance label, rather than the other way around. Under this model, it is possible for a non-normal bag to produce all normal instances, which is disallowed in other MIL models. However, for our system, this is entirely appropriate; it is possible (though unlikely) for a muscle with a myopathic or neurogenic disorder to produce all normal MUPTs. Among existing models, that of Yang et al. [2009] is most similar to BIF.

### 4.1.2 FIB: $B \leftarrow I \leftarrow \vec{F}\ _m$

FIB represents another way of expressing an MIL model where the instances generate the bag label. Under this structure, the feature vectors are drawn from some $P(\vec{F})$, they then generate the instance labels, which in turn determine the bag label. The probability $P(I|\vec{F})$ can be expressed using any discriminative learner, which is attractive, though we still must use a hidden variable method like EM for training because the instance labels are not observed. In order to make the model fully generative we must also model $P(\vec{F})$ using density estimation. In previous work, for example, EM-DD, this model structure is used (though not made explicit) since if the model does *not* need to be generative (i.e. if we will always condition on $\vec{F}$) then density estimation is not required at all. We will show in Section 4.2 that the well-known EM-DD MIL algorithm [Han et al., 2007] can be implemented using this model structure.

Note that $B$ has all $m$ instances as parents, and $m$ can vary from muscle to muscle, so we must express $P(B|\vec{I})$ to allow different sized joint labellings; this is discussed in Section 4.2. Unfortunately, in our 3-class

setting, this structure suffers from an important drawback: it offers no way of prohibiting infeasible instance label assignments, i.e. assignments where for example $I_1 = 2$ and $I_2 = 3$. In order to have a fully consistent generative FIB model, therefore, we must add an additional possible bag "label" $b = 0$ that has positive probability given infeasible labelings. This does not reflect the generative process of the MUPT data, but we can still use this model structure and condition on the event $B \neq 0$ where necessary. Foulds and Smyth [2011] note that prior knowledge about the frequency of instance labels given bag labels is difficult to incorporate with a directed edge from $I$ to $B$.

### 4.1.3 IBF: $B \leftarrow \boxed{I \rightarrow \vec{F}}_m$

Under this structure, the instances are generated independently according to some $P(I)$, and they subsequently generate both the bag label and the feature vectors. This model structure is essentially the "Multi-Instance Mixture Model" of Foulds and Smyth [2011]. As in the FIB model, we have $m$ directed edges from the $I_j$ to $B$, which causes the same drawbacks described in the FIB model but does *not* give us the additional flexibility of using a discriminative learner for the instance labels. Therefore, we will not consider this structure for our generative model. Foulds and Smyth [2011] observe that the EM-DD algorithm can be expressed using this structure given an appropriate model of $P(\vec{F}|I)$ and a "discriminative learning objective"; this is equivalent to our FIB structure described above.

### 4.1.4 Alternative model BFI: $B \rightarrow \boxed{I \leftarrow \vec{F}}_m$

This model attempts to combine two attractive properties: The ability to use a discriminative model $P(I|\vec{F})$, and the ability to easily assign values for $P(I|B)$. Unfortunately, in this model, $B$ and $\vec{F}$ are dependent given $I$, which we know to be untrue in our problem. In this model, normal instances where $B = 1$ may have a different feature distribution from normal instances where $B = 2$. Since we count on being able to generalize from normal instances in normal bags to normal instances in abnormal bags, this model is not appropriate. Note that the BFI model is essentially a clustering model with $I$ as the cluster label and $B$ acting simply as an additional feature along with $\vec{F}$.

## 4.2 Model Components

Choosing the model structure determines the conditional independence properties of our model but does not specify a form for the various distributions. We discuss some possibilities for components of the model that have different assumptions and inductive biases.

### 4.2.1 $P(B)$ and $P(I|B)$ for the BIF Structure

Since $B$ and $I$ take on a small number of discrete values, a tabular representation is appropriate. As noted earlier, one can impose restrictions on possible values of $I|B$ by clamping appropriate values in the conditional probability table.

### 4.2.2 $P(\vec{F}|I)$ for the BIF Structure

Because we assume a continuous feature space, $P(\vec{F}|I)$ can be modeled using any density estimation method. We discuss some well-known possibilities here.

**Multivariate Gaussian** One simple choice for $P(\vec{F}|I = i)$ is a multivariate Gaussian distribution with mean $\mu_i$ and covariance $\Sigma_i$ for $i \in \{1, ..., t\}$. Depending on the availability of data and desired modelling assumptions, one can restrict $\Sigma_i$ to be diagonal.

**Gaussian Copula with KDE Marginals** A drawback of the multivariate Gaussian approach is that much real-world data is not in fact Gaussian. Since we want our generative model to be as realistic as possible, we propose a copula-based model that is practical to estimate and can fit the observed data more closely.

Sklar's theorem [1959] implies that any multivariate density $g$ with marginal densities $g_1, g_2, ..., g_p$ and marginal cumulative distribution functions (CDFs) $G_1, G_2, ..., G_p$ can be expressed in the form

$$g(\vec{f}) = g_1(\vec{f}_{[1]}) \cdot g_2(\vec{f}_{[2]}) \cdot ... \cdot g_p(\vec{f}_{[p]}) \\ \cdot c(G_1(\vec{f}_{[1]}), G_2(\vec{f}_{[2]}), ..., G_p(\vec{f}_{[p]})) \quad (1)$$

where $c$ is a *copula density* that captures the dependence structure of the feature vector $\vec{F} = (\vec{F}_{[1]}, ..., \vec{F}_{[p]})$. If the elements of $\vec{F}$ are independent, then $c \equiv 1$.

Because they are all one-dimensional, the marginals can be estimated well using Kernel Density Estimation (KDE) Alpaydin [2010] even with a modest amount of data, giving $\hat{g}_k$ and $\hat{G}_k$ for $k = 1...p$. This allows our model to capture non-Gaussian aspects of the data, such as relatively heavy or light tails, skewness, or even multi-modality, thus making it more realistic.

The copula model allows us to achieve more high-fidelity marginals without resorting to an unrealistic independence assumption: we can still capture pairwise dependencies in the data by assuming a parametric form for $c$ and estimating the necessary parameters. We will assume a Gaussian copula, whose parameter is the covariance matrix of $(\Phi^{-1}(G_1(\vec{F}_{[1]})), \Phi^{-1}(G_2(\vec{F}_{[2]})), ..., \Phi^{-1}(G_p(\vec{F}_{[p]})))$ where $\Phi^{-1}$ is the inverse of the standard normal CDF.

This can be estimated by the empirical covariance of $(\Phi^{-1}(\hat{G}_1(\vec{F}_{[1]})), \Phi^{-1}(\hat{G}_2(\vec{F}_{[2]})), ..., \Phi^{-1}(\hat{G}_p(\vec{F}_{[p]})))$ over the observed $\vec{f}|I = i$ in the data. (We estimate a separate copula model for each possible value of $I$.) Other, more flexible copula models are possible; we have elected to use the Gaussian copula for simplicity.

**Kernel Density Estimation** Kernel Density Estimation (KDE) is a non-parametric method that estimates a probability density or distribution function by summing up *kernel* functions placed at every observed data point. We use the most common form of KDE, which uses a Gaussian kernel. To choose the kernel width, we employ the *maximum smoothing principle* [Terrell, 1990] as a simple but effective choice; other more fine-tuned choices are possible. The advantage of KDE is that it is capable of modeling complex marginals *and* complex dependencies among the variables of interest, but it does not always work well in moderate to high dimensions.

### 4.2.3 $P(\vec{F})$ for the FIB Structure

In principle, any of the density estimators proposed for $P(\vec{F}|I)$ could be used here; however, the marginal $P(\vec{F})$ is likely to be multi-modal, so the copula or KDE models may be more appropriate.

### 4.2.4 $P(I|\vec{F})$ for the FIB Structure

Since $I$ has a discrete domain, any classification method that supplies class probabilities can be used to model $P(I|\vec{F})$. We examine four such methods.

**Logistic Regression** This well-known model assumes $P(I = i|\vec{F}) \propto \exp^{\vec{\beta}_{i[0]} + \vec{\beta}_{i[1:p]}^\top \vec{f}}, i < t$. Note that the maximum likelihood estimate of $\beta$ is not unique if the data are linearly separable.

**Support Vector Machines** Although the classic SVM formulation [Cortes and Vapnik, 1995] does not provide conditional class probabilities, subsequent work [Huang et al., 2006, Chang and Lin, 2011] has added this capability. It has the added advantage that in the event of feature separability, we get a large-margin classifier whereas logistic regression would fail to converge. Furthermore, kernelized SVMs allow us to easily create non-linear separators in a feature space.

**K Nearest Neighbours** If the decision boundary between instance labels is believed to be complex and if we have sufficient data, a non-parametric model may be warranted. K nearest neighbours uses the empirical distribution of the instance labels of the K closest feature vectors to $\vec{f}$ to estimate $P(I|\vec{F} = \vec{f})$.

**"Diverse Density"** When bag and feature labels are binary ($t = 2$ where 2 is positive and 1 is negative) we may assume $P(I = 2|\vec{F}) = \exp(-\sum_{k=1}^{p} s_{[k]}^2 (\vec{f}_{[k]} - \vec{w}_{[k]})^2)$. Here, $s$ and $\vec{w}$ are parameters fit by maximum likelihood. Note that this is *not* a gaussian distribution; its conditional distributions are Bernoulli such that $P(I = 2|\vec{F} = \vec{f}) \approx 1$ when $\vec{f}$ is near $\vec{w}$. We infer that the reason for its use in the DD and EM-DD algorithms comes from the assumption that the positive instances were localized in feature space, whereas the negative instances were assumed only to be far from the positive ones; they were explicitly assumed not to form a cluster of their own. Because it requires $t = 2$ we cannot use this model for our MUPT data, but we can use it on the MUSK data (described later) for comparison purposes.

### 4.2.5 $P(B|I)$ for the IBF and FIB Structures

Since $m$ varies from bag to bag, we must express $P(B|I)$ as a function that can take a variable number of parameters. Recall that in this setting, we must allow for the possibility that the joint labelling of $\vec{I}$ is not feasible; we add $b = 0$ to the domain of $B$ to capture this event. We can adhere to the standard MIL assumptions by making $P(B|\vec{I})$ deterministic as follows. For feasible labelings, we set

$$P(B = b|I_1, I_2, ..., I_m) = \begin{cases} 1 & \text{if } b = \max_j I_j \\ 0 & \text{otherwise,} \end{cases} \quad (2)$$

and for infeasible labelings we set

$$P(B = b|I_1, I_2, ..., I_m) = \begin{cases} 1 & \text{if } b = 0 \\ 0 & \text{otherwise.} \end{cases} \quad (3)$$

## 5 Learning and Inference

All of the components described in Section 4.2 have associated off-the-shelf learning algorithms for completely observed data. We must learn our models without ever observing $I$ (though with substantial information about $I$ provided through $B$), so we use a hard-Expectation-Maximization (EM) procedure for simultaneously learning the model parameters and inferring the most likely $I$ given the observed $B$ and $\vec{F}$. This worked well on our MUPT data; the conceptual groundwork we lay here could also be used with sampling-based approaches if desired.

### 5.1 Learning

For learning, we use a "hard-EM" approach [Koller and Friedman, 2009]. We assume access to a collection of $n$ bags of the form $(b, \vec{f}_1, \vec{f}_2, ..., \vec{f}_{m_\nu})_\nu, \nu \in$

$\{1, ..., n\}$ which are independent and identically distributed. Given an initial label assignment to all of the instances in our dataset, our learning method has a straightforward implementation; a sketch is presented in Algorithm 1. We discuss the two main steps.

### 5.1.1 Parameter Estimation

**BIF** For the BIF model, we must estimate $P(B)$, $P(\vec{I}|B)$, and $P(\vec{F}|I)$. The marginal probability $P(B)$ is estimated from observed bag label counts only; it does not change across iterations. Because we assume $P(I|B)$ is the same for all instances in all bags, we pool all the bags together and use the aggregated counts to estimate $P(I|B)$. We may add "pseudo-counts" to this estimate if a dirichlet prior is desired; in our experiments we assume for each bag type that we have seen each compatible instance label once, and each incompatible label zero times. To learn $P(\vec{F}|I)$, again we may pool all of the instances together to learn $t$ density estimates $P(\vec{F}|I=1), P(\vec{F}|I=2), ..., P(\vec{F}|I=t)$.

**FIB** For the FIB model, we must estimate $P(\vec{F})$ and $P(I|\vec{F})$; we assume that $P(B|I)$ is fixed according to the standard MIL definition. To estimate $P(\vec{F})$, we pool all feature vectors together and estimate the necessary parameters. These are completely observed so $P(\vec{F})$ does not change across iterations. To learn $P(I|\vec{F})$, again we pool all of the instances together to learn the conditional distribution using a supervised learning method.

### 5.1.2 Label Updating

To update the labels for each bag given the learned parameters, we must find the most likely instance labels $\vec{i}$ given the observed data, that is, we must compute $\mathrm{argmax}_{\vec{i}} P(\vec{I}=\vec{i}|B=b, \vec{F}_1=\vec{f}_1, ..., \vec{F}_m=\vec{f}_m)$ for each bag.

**BIF** From the conditional independence structure of the BIF model, we have

$$P(\vec{I}=\vec{i}|B=b, \vec{F}_1=\vec{f}_1, ..., \vec{F}_m=\vec{f}_m)$$
$$\propto P(\vec{I}=\vec{i}|B=b)P(\vec{F}_1=\vec{f}_1, ..., \vec{F}_m=\vec{f}_m|\vec{I}=\vec{i}).$$

Since the labels and feature vectors for different instances are independent given the bag label, we have

$$P(\vec{I}=\vec{i}|B=b)P(\vec{F}_1=\vec{f}_1, ..., \vec{F}_m=\vec{f}_m|\vec{I}=\vec{i})$$
$$= \prod_{j=1}^{m} P(I_j=i_j|B=b)P(\vec{F}_j=\vec{f}_j|I_j=i_j),$$

so to maximize the probability of the joint label assignment $\vec{i}$, we may maximize each instance label independently:

**Algorithm 1** Hard EM Algorithm Sketch
  **for all** bags **do** {initialize instance labels}
    $\vec{i} \leftarrow b$
  **end for**
  **repeat**
    learn model components {M-step}
    **for all** bags **do** {relabel instances: E-step}
      $\vec{i} \leftarrow \mathrm{argmax}_{\vec{i}} P(\vec{I}=\vec{i}|B=b,$
      $\qquad\qquad\qquad \vec{F}_1=\vec{f}_1, ..., \vec{F}_m=\vec{f}_m)$
    **end for**
  **until** instance labels do not change

$$\vec{i}^*_{[j]} \leftarrow \mathrm{argmax}_{i_j \in \{1,...,t\}} P(I_j=i_j|B=b)P(\vec{F}_j=\vec{f}_j|I_j=i_j).$$

**FIB** From the conditional independence structure of the FIB model, we have

$$P(\vec{I}=\vec{i}|B=b, \vec{F}_1=\vec{f}_1, ..., \vec{F}_m=\vec{f}_m)$$
$$\propto P(B=b|\vec{I}=\vec{i})P(\vec{I}=\vec{i}|\vec{F}_1=\vec{f}_1, ..., \vec{F}_m=\vec{f}_m).$$

However, in this model, the instance labels are *not* conditionally independent given the bag label and we cannot maximize them independently. For example, if $B=2$, and $I_1, I_2, ..., I_{m-1}$ are all equal to 1, then $I_m$ must be equal to 2 according to the MIL assumption encoded in $P(B|\vec{I})$. However, we can still avoid searching over all $t^m$ possible label vectors. We defined in Section 4 that if $\vec{i}$ is a feasible vector for $B=b$, then we have $P(B=b|\vec{i})=1$ and therefore we have

$$P(B=b|\vec{i})P(\vec{I}=\vec{i}|\vec{F}=\vec{f}) = \prod_{j=1}^{m} P(I_j=i_j|\vec{F}_j=\vec{f}_j).$$

We can find the best feasible $\vec{i}$ in two steps:

1. Let $\vec{i}^*_{[j]} \leftarrow \mathrm{argmax}_{i_j \in \{1\} \cup \{b\}} P(I_j=i_j|\vec{F}_j=\vec{f}_j)$

2. If $\vec{i}^* = \vec{1}$, set $\vec{i}^*_{[k^*]} \leftarrow b$ for $k^*$ given by
$$\mathrm{argmax}_{k} \left[ P(I_k=b|\vec{F}_j=\vec{f}_j) \prod_{j \neq k} P(I_j=1|\vec{F}_j=\vec{f}_j) \right].$$

The vector $\vec{i}^*$ is feasible and maximizes $\prod_{j=1}^{m} P(I_j=i_j|\vec{F}_j=\vec{f}_j)$ over all feasible vectors when $B=b$.

### 5.2 Inference

Once we have learned all of the model parameters, given a new bag where only feature values $\vec{f}$ are observed, we wish to compute $\mathrm{argmax}_{\vec{i},b} P(I=\vec{i}, B=b|\vec{F}_1=\vec{f}_1, ..., \vec{F}_m=\vec{f}_m)$. These are the most likely bag and instance labels given the $m$ feature vectors in the new bag.

**BIF**  In the BIF model,

$$\operatorname*{argmax}_{\vec{i},b} P(\vec{I}=\vec{i}, B=b|\vec{F}_1=\vec{f}_1,...,\vec{F}_m=\vec{f}_m)$$
$$= \operatorname*{argmax}_{\vec{i},b} P(B=b)P(\vec{I}=\vec{i}|B=b)$$
$$\cdot P(\vec{F}_1=\vec{f}_1,...,\vec{F}_m=\vec{f}_m|\vec{I}=\vec{i})$$
$$= \operatorname*{argmax}_{b}\left[P(B=b)\operatorname*{argmax}_{\vec{i}}\left(P(\vec{I}=\vec{i}|B=b)\right.\right.$$
$$\left.\left.\cdot P(\vec{F}_1=\vec{f}_1,...,\vec{F}_m=\vec{f}_m|\vec{I}=\vec{i})\right)\right].$$

Therefore we can apply the instance label updating method presented in Section 5.1 for each possible bag label and weight them according to $P(B)$ to find the joint MAP assignment to $b$ and $\vec{i}$.

**FIB**  In the FIB model,

$$\operatorname*{argmax}_{\vec{i},b} P(\vec{I}=\vec{i}, B=b|\vec{F}_1=\vec{f}_1,...,\vec{F}_m=\vec{f}_m)$$
$$= \operatorname*{argmax}_{b}\left[\operatorname*{argmax}_{\vec{i}}\left(P(B=b|\vec{I}=\vec{i})\right.\right.$$
$$\left.\left.\cdot P(\vec{I}=\vec{i}|\vec{F}_1=\vec{f}_1,...,\vec{F}_m=\vec{f}_m)\right)\right]$$

Therefore we can apply the instance label updating method presented in Section 5.1 for each possible bag label to find the joint MAP assignment to $b$ and $\vec{i}$.

## 6  Experiments

We now give the details of how our MUPT dataset was constructed, and we discuss the results of the various generative models as applied to our MUPT dataset. We find that the BIF structured models perform very well for several different component choices. The FIB structured models perform less well, but still much better than chance on both bags and instances. Based on our results, we recommend structure and component choices that lead to a high-fidelity generative model of MUPT data. We also give the performance of the models on the MUSK1 dataset [Dietterich et al., 1997].

### 6.1  The MUPT dataset

Recall that each detected EMG signal is a composite signal that represents the activity of all of the MUs that were active during a muscle contraction. After acquiring an EMG signal, it is decomposed into its constituent MUPTs, each of which ideally represents the electrical activity of a single, sampled MU. As such, the MUPTs are our instances and are the source of our instance level features. From each MUPT, it is common practice to compute a *MUP template* which is a single MUP whose shape is representative of all MUPs in the MUPT [Stashuk, 1999]. All but two of the features we use are functions of this MUP template, while the rest of the features describe aspects of the MUPT itself. We use $p=8$ features, which were chosen by automated feature selection (both wrapper- and filter-based) in prior work [Adel et al., 2012].

1. **Number of turns** is the number of positive and negative peaks; a function of the MUP template.
2. **Amplitude** represents the maximum difference of voltage between two points [Dumitru et al., 1995]; a function of the MUP template.
3. **Area** represents the area under the curve; a function of the MUP template.
4. **Thickness** refers to the ratio of area to amplitude; a function of the MUP template.
5. **Size index** given by $2\log(\text{amplitude}) + \frac{\text{area}}{\text{amplitude}}$; a function of the MUP template.
6. **Turn width** is given by $\frac{\text{duration}}{\text{turns}}$. Duration is the interval from the first signal deflection from baseline to its final return to baseline [Dumitru et al., 1995]; a function of the MUP template.
7. **Firing rate MCD (mean consecutive difference)** refers to the sequential change in the firing rate of the MU over time; a function of the MUPT.
8. **A-jiggle** is a measure of the shape variability of band-pass filtered MUPs ($2^{\text{nd}}$ derivative of the signal); a function of the MUPT.

We have two MUPT datasets, one acquired from upper-leg recordings containing 88 bags and 1534 instances, and another acquired from lower-leg recordings with 70 bags and 1500 instances. All data were collected under IRB approval and and were de-identified. Prior versions of the MUPT data were used by Adel et al. [2012]; our versions have been cleaned to remove obvious outlier errors. For example, instances with highly improbable feature values were removed.

### 6.2  Results

Table 1 shows the performance of different models on our data. Because one of the authors [TA] manually labeled the instances, we can estimate both the accuracy of each model for classifying bags *and* the accuracy for classifying instances given only the features within a new bag. Note, the manually assigned instance labels were *not* used for learning or inference. The accuracy results were computed using leave-one-bag-out cross-validation. We also present the log likelihood of the observed data maximized over the model parameters and the hidden instance labels, which measures how well the models fit the training data.

We present the results for the BIF structure using five different density estimators for $P(\vec{F}|I)$. We use two versions each of the Gaussian and copula models, one assuming independence between elements of the feature vectors given the instance labels (e.g. a diagonal covariance matrix) indicated by the prefix $\perp\!\!\!\perp$, and one assuming pairwise correlations. The marginals of the copula models are estimated using KDE with a Gaussian kernel and the maximum smoothing principle (MSP) bandwidth [Terrell, 1990]. We also give results for a multi-dimensional KDE for $P(\vec{F}|I)$, again with the MSP bandwidth.

Table 1: MUPT Dataset Results. To give a sense of the statistical uncertainty, we mark all accuracies that are within the 99% Bernoulli confidence interval of the maximum observed accuracy in bold. We mark the highest log likelihoods for the BIF and FIB structures in italics.

| Upper Leg | BIF: $B \rightarrow I \rightarrow \vec{F}_m$ | | | | | FIB: $B \leftarrow I \leftarrow \vec{F}_m$ | | | | Non-MIL |
|---|---|---|---|---|---|---|---|---|---|---|
| Rnd: 0.33 | $\perp\!\!\!\perp$Gauss | $\perp\!\!\!\perp$Cop. | Gauss | Cop. | KDE | LR | KNN | SVM | QDA | $\perp\!\!\!\perp$QDA |
| Bag Acc. | **0.955** | **0.955** | **0.955** | **0.955** | **0.955** | 0.728 | 0.568 | 0.250 | **0.898** | 0.841 |
| Inst. Acc. | **0.984** | **0.978** | **0.983** | **0.980** | **0.983** | 0.728 | 0.674 | 0.415 | **0.978** | 0.850 |
| Log lik. | -36843 | -37104 | -36810 | -37066 | *-34726* | *-32889* | -32998 | -34382 | -32938 | — |
| Lower Leg | BIF: $B \rightarrow I \rightarrow \vec{F}_m$ | | | | | FIB: $B \leftarrow I \leftarrow \vec{F}_m$ | | | | Non-MIL |
| Rnd: 0.33 | $\perp\!\!\!\perp$Gauss | $\perp\!\!\!\perp$Cop. | Gauss | Cop. | KDE | LR | KNN | SVM | QDA | $\perp\!\!\!\perp$QDA |
| Bag Acc. | **0.986** | **0.971** | **0.971** | **0.957** | 0.886 | 0.814 | 0.586 | 0.371 | 0.886 | 0.771 |
| Inst. Acc. | **0.946** | 0.899 | **0.931** | 0.880 | 0.859 | 0.543 | 0.571 | 0.469 | 0.915 | 0.781 |
| Log lik. | -38035 | -38206 | -37980 | -38141 | *-35999* | *-34833* | -34952 | -35598 | -35128 | — |

Table 2: MUSK1 Dataset Results

| MUSK1 | BIF: $B \rightarrow I \rightarrow \vec{F}_m$ | | | | | FIB: $B \leftarrow I \leftarrow \vec{F}_m$ | | | | | Non-MIL |
|---|---|---|---|---|---|---|---|---|---|---|---|
| Rnd: 0.50 | $\perp\!\!\!\perp$ Gauss | $\perp\!\!\!\perp$ Cop. | Gauss | Cop. | KDE | LR | KNN | SVM | QDA | DD | QDA |
| Bag Acc. | **0.870** | **0.848** | 0.696 | 0.641 | **0.772** | **0.783** | **0.772** | **0.837** | **0.837** | 0.620 | **.783** |
| Log lik. | *-14921* | -18437 | -45815 | -51031 | -33591 | -34086 | -34120 | -34076 | *-34056* | -34105 | — |

We present results for the FIB structure using four different discriminative learning models. In all cases, $P(\vec{F})$ was estimated using a multi-dimensional KDE with the MSP bandwidth. The discriminative learners were Logistic Regression (LR), K-nearest neighbors with $K = 7$ (KNN), SVMs with a radial basis function kernel, $C = 1$ and $\gamma = 1/8$, and Quadratic Discriminant Analysis (QDA). The parameter $K$ was chosen based on past experience with the data; SVM parameters are defaults. In the last column, we also present results using Quadratic Discriminant Analysis in a *non* multiple-instance setting by assuming the instance labels are in fact the bag labels and labelling new bags by majority vote.

Table 2 shows results on the MUSK1 dataset, which contains 92 bags and 476 instances. We use the same models and add a version of the FIB model with the "Diverse Density" (DD) model for $P(I|\vec{F})$. Since the data are 166-dimensional, as a pre-processing step, we use PCA to eliminate near-collinearity; we choose enough components to capture 90% of the variance, leaving us with $p = 76$ features. Results are not state-of-the-art—Zhang and Goldman [2001] achieve 96.8%—but moderately good; among our models the BIF model with independent Gaussians for $P(\vec{F}|I)$ has the highest cross-validation accuracy and log likelihood. No expert instance labels exist for MUSK1.

## 7 Conclusions

Results on the MUPT data indicate that all of our BIF-based generative models perform better than previous state-of-the-art work by Adel et al. [2012], whose best leave-one-bag-out bag label accuracy was 82.3% (lower leg.) In addition, we demonstrate that we are able to recover the instance labels with very high accuracy. The FIB models had worse performance on the MUPT data but better on the MUSK1 data, suggesting they may be useful for other tasks. If muscle classification accuracy is paramount, the parametric model components (Gaussian and Copula) appear best, but if high-fidelity simulation is paramount, then the KDE model component is a better fit to the observed data.

We have introduced a general framework for generative models in MIL. Although MIL is a well-developed sub-field of Machine Learning, generative model approaches had not received much attention so far. Our results suggest that models that are well aligned with the actual data generation in a problem domain (the BIF structure in the case of our muscle classification task) are an excellent choice for classification and modeling purposes.